\documentclass[letterpaper, 10 pt, conference]{ieeeconf}  

\IEEEoverridecommandlockouts                              

\usepackage{graphics} 
\usepackage{epsfig} 
\usepackage{graphicx}
\usepackage{amsmath}
\usepackage{algorithm}
\usepackage{algpseudocode}
\usepackage{svg}
\usepackage{multirow}
\usepackage{multicol}
\usepackage{booktabs,subcaption,amsfonts,dcolumn}
\usepackage{tablefootnote}

\usepackage{tikz}

\newcommand\submittedtext{%
  \footnotesize This work has been submitted to the IEEE for possible publication. Copyright may be transferred without notice, after which this version may no longer be accessible.}

\newcommand\submittednotice{%
\begin{tikzpicture}[remember picture,overlay]
\node[anchor=south,yshift=10pt] at (current page.south) {\fbox{\parbox{\dimexpr0.65\textwidth-\fboxsep-\fboxrule\relax}{\submittedtext}}};
\end{tikzpicture}%
}

\title{\LARGE \bf
Self-Organizing Edge Computing Distribution Framework \\for Visual SLAM
}

\author{Jussi Kalliola$^{1*}$, Lauri Suomela$^{1}$, Sergio Moreschini$^{1}$, David Hästbacka$^{1}$
\thanks{$^*$ Corresponding Author 
}
\thanks{This work is part of the AeroPolis project funded by the Academy of Finland (grant number 348481).}%
\thanks{$^{1}$Tampere university, Tampere, Finland. \{jussi.kalliola, lauri.a.suomela, sergio.moreschini, david.hastbacka\}@tuni.fi}
}

\begin{document}

\maketitle
\submittednotice
\thispagestyle{empty}
\pagestyle{empty}

\begin{abstract}
Localization within a known environment is a crucial capability for mobile robots.
Simultaneous Localization and Mapping (SLAM) is a prominent solution to this problem.
SLAM is a framework that consists of a diverse set of computational tasks ranging from real-time tracking to computation-intensive map optimization.
This combination can present a challenge for resource-limited mobile robots. 
Previously, edge-assisted SLAM methods have
demonstrated promising real-time execution capabilities
by offloading heavy computations while performing real-time tracking onboard. 
However, the common approach of utilizing a client-server architecture for offloading is sensitive to server and network failures.
In this article, we propose a novel edge-assisted SLAM framework capable of self-organizing fully distributed SLAM execution across a network of devices or functioning on a single device without connectivity.
The architecture consists of three layers and is designed to be device-agnostic, resilient to network failures, and minimally invasive to the core SLAM system. 
We have implemented and demonstrated the framework for monocular ORB SLAM3 
and evaluated it in both fully distributed and standalone SLAM configurations against the ORB SLAM3.
The experiment results demonstrate that the proposed design matches the accuracy and resource utilization of the monolithic approach while enabling collaborative execution.

\end{abstract}

\section{INTRODUCTION}

Autonomous mobile robots are increasingly being deployed in applications like manufacturing, search-and-rescue, and logistics, but face challenges with limited resources for complex tasks requiring advanced computation and sensory input. 
The race to address increasingly complex tasks has accelerated the demand for improved computational capabilities; however, increasing computational resources is not always feasible due to associated increases in energy consumption which limit operational time. 
As a result, computational offloading to cloud or edge environments has garnered interest in utilizing resource-rich environments to meet computational requirements.

\begin{figure}
  \centering
  \includegraphics[width=\linewidth]{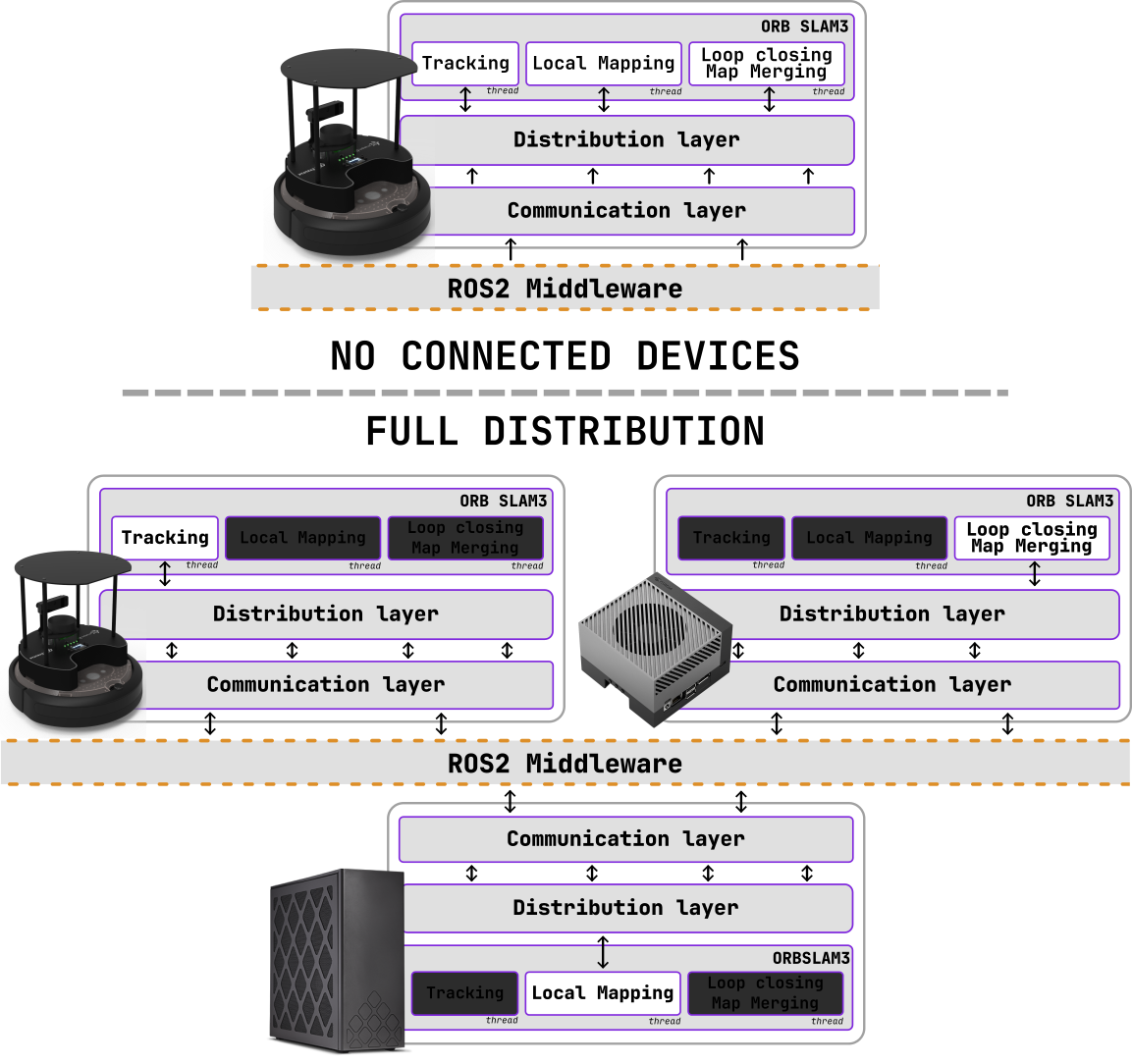}
  \caption{High-level illustration of the proposed distribution framework for VSLAM. 
  The system in multi-node (bottom) and single-node configuration (top). 
  }
\end{figure}

A critical function for mobile robots is the ability to map and localize within their environment using onboard sensors, known as Simultaneous Localization and Mapping (SLAM). 
Although SLAM methods have evolved, including extensions to multi-robot systems \cite{covins_schmuck_21, ccslam_schmuck_19, dcslam_zhang_18}, they continue to demand significant computational resources to achieve high accuracy in real-time. 
Efforts to mitigate these demands include improvements in areas such as algorithmic development \cite{slam_midterm_burgard_08, orbslam3_campos_21, vinsmono_li_18, lif_slam_dl_bruno_21}, and computational offloading to edge or cloud \cite{ali_edgeslam_22, swarmmap_xu_22, edgeslam2_li_24}. 
This article focuses on the development of edge-assisted SLAM methods and proposes a novel framework for distributed Visual SLAM capable of self-organizing fully distributed SLAM execution over the network of devices.


Several frameworks have been proposed to offload computationally intensive map optimization to an edge server while executing time-sensitive tracking onboard \cite{edgeslam_semantic_cao_23, ali_edgeslam_22, swarmmap_xu_22, edgeslam2_li_24}. 
These approaches have demonstrated impressive results in areas such as reduced CPU usage \cite{dynnetslam_sossala_22, cloudslam_wright_20, ali_edgeslam_22} and improved energy efficiency \cite{edge_slam_energy_sossala_23}. 
However, previous methods typically rely on client-server architectures that are designed to offload only specific tasks(s) to the server \cite{ali_edgeslam_22, edgeslam_semantic_cao_23, ccslam_schmuck_19, cloudslam_wright_20, edgeslam2_li_24}.
While promising, we have identified some limitations:
1) Systems are designed to offload only specific module(s), making adjustments labor-intensive;
2) They are vulnerable to server or communication failures, relying solely on one server.


To address these limitations, we introduce a novel self-organizing distribution framework for Visual SLAM (VSLAM), which can fully distribute all SLAM modules across nodes in a network. 
Self-organizing distributed SLAM execution is orchestrated by a heuristic-based distribution policy which enables single-node or multi-node SLAM execution.
In a multi-node configuration, nodes self-organize distributed SLAM execution, and in a single-node setup, the system reverts to standalone SLAM. 
Redundancy introduced by a shared two-tier system state enhances resiliency to network and node failures since other nodes can fulfill lost resources.


We demonstrate the proposed framework by implementing it as a minimally invasive wrapper around monocular ORB SLAM3. 
The architecture comprises three layers: core, distribution, and communication. 
The core layer incorporates monocular ORB SLAM3 with minimal adjustments. 
The distribution layer intermediates between the core and communication layers, using a distribution policy to orchestrate distributed SLAM execution. 
Additionally, this layer manages the system state to ensure eventual consistency across nodes. 
The communication layer leverages ROS2 and FastDDS middleware to facilitate interaction with other network nodes. 
The system is designed to be device-agnostic, the architecture supports self-connection and synchronization across multiple devices.


We evaluated the framework using benchmark datasets from handheld sensors and robot platforms alongside real-world experiments. 
The system’s accuracy and resource utilization were validated against ORB SLAM3. 
Results demonstrate that SLAM modules can be successfully distributed across heterogeneous devices over a network, achieving comparable Average Trajectory Error (ATE) and CPU utilization in most of the experiments. 
The system shows marginally lower CPU usage, although at the expense of network usage.
Additionally, experiments show that the system can revert to standalone SLAM in the case of single-device configuration.
The added complexity of the distributed system introduces certain limitations; in some experiments, minor state management failures were observed, leading to a reduction in accuracy.


We summarize our contributions as follows: \begin{enumerate} 
\item We propose a reconceptualized edge-assisted distributed SLAM framework enabling full distribution, device-agnosticity, and resilience to network failures.
\item A novel distribution layer with a heuristic-based distribution policy enables self-organizing collaborative SLAM execution across heterogeneous networked devices.
\item Finally, we have implemented and demonstrated the distribution framework for monocular ORB SLAM3. The distributed system achieves comparable ATE and resource utilization to the original system in single and multi-node configurations.
\end{enumerate}


\section{RELATED WORK}

\subsection{Visual SLAM}

SLAM methods have been studied since the 1980s \cite{seminal_slam_lemmer_88, seminal_slam_smith_86} and have evolved to support various sensors \cite{orbslam3_campos_21}, and extended to multi-robot teams \cite{ccslam_schmuck_19, covinsg_patel_23}. 
In this article, we focus specifically on monocular VSLAM. MonoSLAM was one of the first proposed solutions for monocular VSLAM, using an Extended Kalman Filter (EKF) and Shi-Tomasi points as tracked features  \cite{monoslam_davison_03_1, monoslam_davison_07_2, monoslam_civera_08_3}. 
Over the years, several feature extraction methods have been developed for feature tracking, spanning from traditional computer vision techniques \cite{orbslam1_murartal_15, orbslam2_murartal_16, orbslam3_campos_21, cudasift_slam_tardos_24, vinsmono_li_18} to modern deep learning approaches \cite{lif_slam_dl_bruno_21}. 
To ensure consistency of matched features, various data association methods have been introduced \cite{slam_midterm_burgard_08, ransac_civera_10}, including more recent mid-term data association methods based on deep learning \cite{droid_teed_21}.
Recently, keyframe-based approaches have become popular due to the reduced computational requirements with Bundle Adjustment (BA) optimization \cite{orbslam3_campos_21}. 
These approaches use a sparse set of keyframes for the map estimation instead of all frames captured by the sensor.
Long-term data association enables relocalization, loop closing, merging disconnected maps, and global map optimization in SLAM systems \cite{orbslam3_campos_21}.
The bag-of-words library DBoW2 is a common solution for long-term associations \cite{dbow_galvez_12}, which is used in VINSMono \cite{vinsmono_li_18} and ORB SLAM3 \cite{orbslam3_campos_21}. 
Many SLAM approaches have adopted parallel execution by separating functionality into two or more threads \cite{ptam_klein_07, orbslam3_campos_21, orbslam2_murartal_16, vinsmono_li_18, droid_teed_21}. 
One approach is to separate short-term (tracking), mid-term (Local BA), and long-term (loop closing) data associations into their own threads \cite{orbslam3_campos_21}.

\subsection{Collaborative SLAM}

Collaborative SLAM (CSLAM) methods can be categorized into centralized or distributed architectures. 
CCM-SLAM \cite{ccslam_schmuck_19} utilizes centralized architecture, where a central server manages maps for a team of robots, for visual-inertial CSLAM. 
COVINS \cite{covins_schmuck_21, covinsg_patel_23} builds on CCM-SLAM \cite{ccslam_schmuck_19} and is capable of scaling up to 12 robots.
In distributed architectures, the centralized server is removed, and robots collaborate through local connections. 
One of the first distributed monocular CSLAM methods was proposed where robots performed global map merging onboard \cite{dcslam_zhang_18}. 
Later work has focused on distributed Pose Graph Optimization (PGO), proposing methods such as Distributed Gauss-Seidel (DGS) \cite{dgs_choudhary_17}. 

This brief overview of CSLAM methods establishes a distinction between our work and previous CSLAM research. 
Our focus is on \textit{software framework for distributed VSLAM}, while CSLAM primarily emphasizes algorithmic development such as distributed PGO for multi-robot systems. 
Therefore, our work aligns more closely with \textit{Edge-assisted SLAM}, which is covered next.



\subsection{Edge-assisted SLAM}

In edge-assisted SLAM research, the goal is to enable real-time execution on platforms with limited resources by offloading computations from mobile robots to edge servers.
A common approach is to offload computationally intensive tasks, such as map optimization while executing time-sensitive tracking on the mobile device \cite{edgeslam_semantic_cao_23, ali_edgeslam_22, swarmmap_xu_22, edgeslam2_li_24}.
An alternative solution, AdaptSLAM, was proposed to run tightly coupled Tracking (TR) and Local Mapping (LM) on a mobile device while offloading Loop Closing (LC) and Global Bundle Adjustment (GBA) to an edge server \cite{adapt_slam_chen_23}. 
A standard approach is to use a client-server architecture for offloading tasks from the mobile device to an edge server \cite{ali_edgeslam_22, edgeslam2_li_24, edgeslam_semantic_cao_23, ccslam_schmuck_19, edgeslam2_li_24, adapt_slam_chen_23}.
Additionally, proposed solutions often involve extensive modifications to existing SLAM methods \cite{cloudslam_wright_20, swarmmap_xu_22, edgeslam2_li_24}, which makes it difficult to abstract the design choices and implement offloading schemes to other systems.
Resource utilization and optimization have also been core topics in edge-assisted SLAM research. 
Network bandwidth is a widely recognized metric for evaluating performance \cite{ali_edgeslam_22, cloudslam_wright_20, swarmmap_xu_22, ccslam_schmuck_19}, along with energy consumption \cite{edge_slam_energy_sossala_23, edgeslam2_li_24}, CPU usage \cite{ali_edgeslam_22}, and memory usage \cite{swarmmap_xu_22, ali_edgeslam_22}.
A recent study proposed EdgeSlam2 which integrates
TR and LM on mobile devices using a System-on-Chip (SoC) and implementing a lightweight Loop Closure on the mobile device \cite{edgeslam2_li_24}.

\section{DISTRIBUTED VSLAM FRAMEWORK}

This section outlines the framework for distributed VSLAM, focusing on key concepts including initialization, connectivity, state management, distribution policy, and requirements. We assume that the core SLAM system is divided into three parallel threads, each dedicated to executing short-term, mid-term, and long-term data associations, respectively.



\subsection{Initialization} 
\label{subsect:init}

A uniform system is implemented across all compute nodes, denoted as \(N^{tr}\) for Tracking (TR), \(N^{lm}\) for Local Mapping (LM), and \(N^{lc}\) for Loop Closing (LC). A set of compute nodes is denoted as \(C=\{N^{tr}, N^{lm}, N^{lc}\}\).
During initialization, each node is assigned a unique task ID, which defines its corresponding task.
In the discovery phase, connectivity is established by broadcasting the task ID to other nodes in the network. 
Additionally, other nodes adjust their distribution policy based on the discovery phase. 
In parallel, the SLAM system is initialized.

A common approach in VSLAM systems for map initialization involves creating an initial map \(M^1 = \{ K^1, P^1 \}\) from two consecutive frames, denoted as keyframes \(K^1 = \{K_1, K_2\}\), and the initial set of 3D map points \(P^1\) is computed.
In the distributed system, these are transmitted to \(N^{lm}\), where poses are optimized.
However, during the optimization, \(N^{tr}\) still uses the unoptimized KFs for tracking which is prone to failure.
Therefore, the distributed SLAM system needs to wait that initial KFs are optimized before determining that tracking is lost. 


\subsection{Connectivity} 

\begin{figure}[tp]
  \centering
  \smallskip
  \smallskip
  \includegraphics[width=0.85\columnwidth]{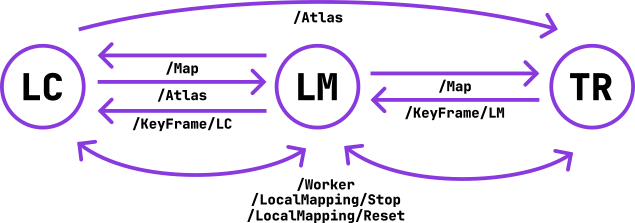}
  \caption{Connectivity scheme. Arrows indicate the direction of the data flow and its corresponding ROS2 topic.}
  \label{fig:conn_scheme}
\end{figure}

Figure~\ref{fig:conn_scheme} illustrates the proposed connectivity scheme, showing the direction and topic of the data flow. 
As shown, 
\(N^{lm}\) functions as a gateway between 
\(N^{lc}\) and 
\(N^{tr}\), as publishing data to multiple destinations from \(N^{tr}\) can introduce additional latency. 

In general, distributing computations over the network introduces data transmission and processing latencies. 
Specifically, the total latency from \(N^i\) to \(N^j\) is expressed by \(L(i,j)=T_p(i,j) + T_{proc}(i,j) + T_{proc}(j,i) + T_{p}(j,i)\), where \(T_p()\) is the transmission latency and \(T_{proc}()\) processing latency. 
In practice, both processing and transmission latency may vary depending on the direction of the data flow.


\subsection{State management model} 

Given that nodes in the network do not share the same physical memory, the system's successful execution requires each node to maintain a copy of the system state. 
Ideally, all nodes would maintain an identical state at all times, enabling the system to operate as the core SLAM system.
However, in practice, the state across nodes may vary or be incomplete because of the additional latency. 
Thus, state management becomes a critical component of the framework. 

State refers to a set of disconnected maps \(S = \{ M^1, M^2, \dots, M^n \}\).
We propose to handle distributed state management by introducing two distinct types of state: the overall state \(S_{full}\) and the SLAM state \(S_{slam}\). At a node \(N^i\), the overall state \(S^i_{full}\) is a superset of \(S^{i}_{slam}\), implying that \(S^i_{full}\) includes \(S^{i}_{slam}\) components, such as keyframes \(K^{i}_{slam}\) and Map Points \(P^{i}_{slam}\), as well as components from other nodes in a set \(C\). State of \(N^j\) is expressed as follows \(S^{j}_{full}=\{\{K^j_1, P^j_1\}, \{K^j_2, P^j_2\}, \dots, \{K^j_n, P^j_n\}\}\). 
Thus, the overall state of \(N^i\) is defined as a merged state derived from the states of all other nodes: \(S^i_{full}  = \left\{ \bigcup_{k=j}^{n} K^k, \bigcup_{k=j}^{n} P^k \right\} \), where \(i \in [j,n]\).

In practice, \(S^i_{full}\) contains only a partial \(S^{j_p}_{full} \subset S^{j}_{full}\) state from other nodes at the time \(t\) due to latency and communication between nodes. 
This introduces inconsistency across nodes, which is the reason for two distinct states; the responsibility of \(S_{full}\) is to guarantee that updates to \(S_{slam}\) are complete. 
Incomplete updates can produce an incorrect \(S_{slam}\), which would propagate errors in future optimization and estimation steps, resulting in poor performance. 
Additionally, since \(S_{full}\) cannot guarantee complete state representation from all nodes in the network at every timestep \(t\), an eventual consistency model \cite{distributed_steen_2023} from distributed systems is adopted.
Nodes are effectively available and in soft-state, until the system has functioned long enough to achieve consistency \cite{distributed_steen_2023}.
To maximize consistency across nodes, we propose a simple distribution policy to orchestrate the distribution system.

\subsection{Distribution Policy} 

The system distributes computations and data using a simple heuristic-based distribution policy based on discovered devices in the network. 
The nodes are discovered in the network during the initialization phase. The discovered nodes are expressed as a set $G$. 
The distribution policy can be expressed as follows: If \(N^{lm} \in G\), then \(N^{tr} \rightarrow N^{lm} \) (offload to \(N^{lm}\)), otherwise, perform LM at \(N^{tr}\). 
Similarly, if \(N^{lc} \in G\), then \(N^{tr} \rightarrow N^{lc} \) (offload to \(N^{lc}\)), otherwise, perform LC at \(N^{tr}\).
Thus, the final distribution policy function \( D(N, G) \) at \( N^{tr} \) can be summarized as:
\[
\resizebox{\columnwidth}{!}{$
D(N, G) = \begin{cases} 
      N^{lm} \in G \text{ and } N^{lc} \in G & \rightarrow \text{offload LM to } N^{lm} \text{ and LC to } N^{lc}, \\
      N^{lm} \in G \text{ and } N^{lc} \notin G & \rightarrow \text{offload LM and LC to } N^{lm}, \\
      N^{lm} \notin G \text{ and } N^{lc} \in G & \rightarrow \text{perform LM locally, offload LC to } N^{lc}, \\
      N^{lm} \notin G \text{ and } N^{lc} \notin G & \rightarrow \text{perform all tasks locally}.
   \end{cases}
$}
\] 


The same distribution policy applies to other nodes, following the connectivity scheme illustrated in Figure~\ref{fig:conn_scheme}. 
We have defined two computational offloading tasks and three data distribution tasks. 
The computational tasks are the SLAM modules. 
The data distribution tasks include publishing new KeyFrame(s), Local Map(s), and Global Map(s). 
The data distribution message carries the computational offloading signal. 
Hence, there is no delay or orchestration between data distribution (state update) and computation offloading (collaborative SLAM execution).


\subsection{Requirements for SLAM distribution} 
\label{requirements}

The distribution system must ensure both real-time processing and eventual consistency across nodes. 
Real-time processing is crucial since VSLAM is particularly sensitive to delays.
Eventual consistency is equally important to ensure the reliability of data across all nodes. 
Specifically, the state \( S_{\text{full}} \) at all nodes must remain consistent within a time interval \( t_k \), which can be expressed as:
\(
S_{\text{full}}^i(t) = S_{\text{full}}^j(t) \quad \forall \, i, j \in C, \; t \in t_k
\).
Furthermore, failure to meet even a single requirement may compromise the system and yield suboptimal results. 
Consequently, we deem the requirements fulfilled, and distribution successful, if the system demonstrates comparable accuracy, maintains real-time processing, and achieves eventual consistency across nodes.

\section{DISTRIBUTED MONOCULAR ORB SLAM3} \label{sect:architecture}

We have implemented and demonstrated the framework for monocular ORB SLAM3. The implementation is designed as a three-layer architecture: the first layer contains the SLAM system \textbf{(core)}, the second layer handles data distribution and state management \textbf{(distribution layer)}, and the third layer interacts with the communication middleware \textbf{(communication layer)}.


\subsection{Core: ORB SLAM3}
\label{subsect:core}

Monocular ORB SLAM3 \cite{orbslam3_campos_21} was selected as the core SLAM system because it is decomposed into separate threads by the authors and the monocular system has a high real-time processing requirement.
Additionally, ORB SLAM methods have been well-established in edge-assisted SLAM research \cite{ali_edgeslam_22, swarmmap_xu_22, adapt_slam_chen_23, edgeslam2_li_24} serving as core systems.
The distribution system surrounding the core is designed to be minimally invasive, allowing the design choices to be adapted to other SLAM systems.

The tracking module was modified to reduce the rate of keyframe construction. Keyframes are now created only if at least two frames have elapsed since the last keyframe, and the reference map point ratio was adjusted from 90\% to 80\%, requiring a higher number of tracked points for keyframe generation.
Additionally, initialization process was adjusted to wait until the initial keyframes are optimized before establishing a new map in the event of tracking loss. Without this adjustment, the system would often start a new map immediately. 
Finally, Map Point identifiers were changed from integers to hashed strings to facilitate simultaneous generation across multiple nodes. 
Other modifications were related to memory management, class constructions, and class parameters.



\begin{figure}[tp]
  \centering
  \smallskip
  \smallskip
    \includegraphics[width=0.9\columnwidth]{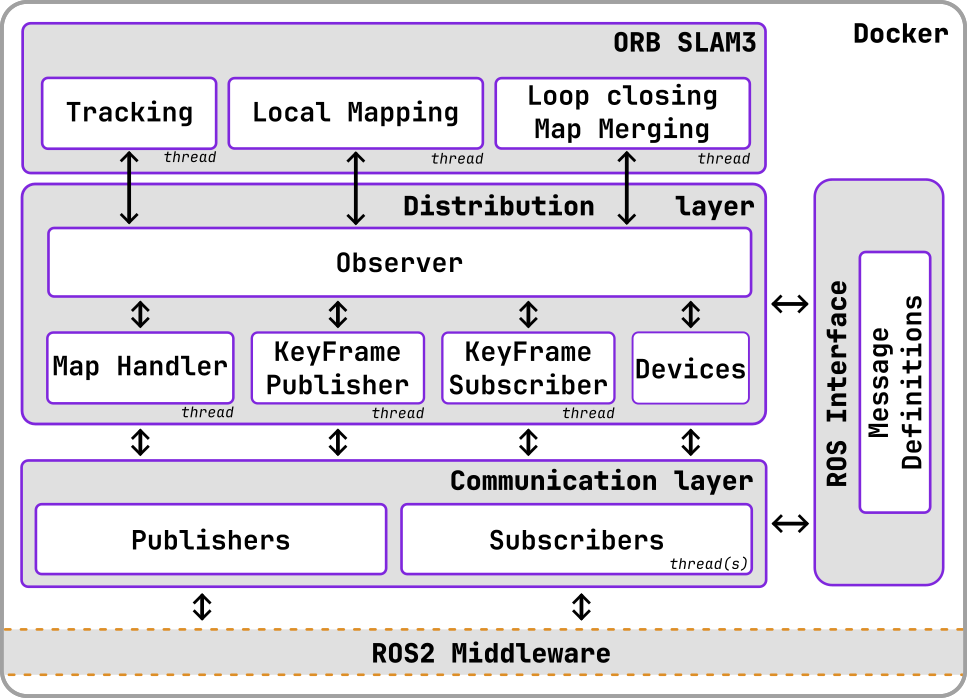}
  \caption{Distribution system architecture.}
  \label{fig:sys_arch}
\end{figure}

\begin{figure*}[tp]
  \smallskip
  \smallskip
    \includegraphics[width=0.9\linewidth]{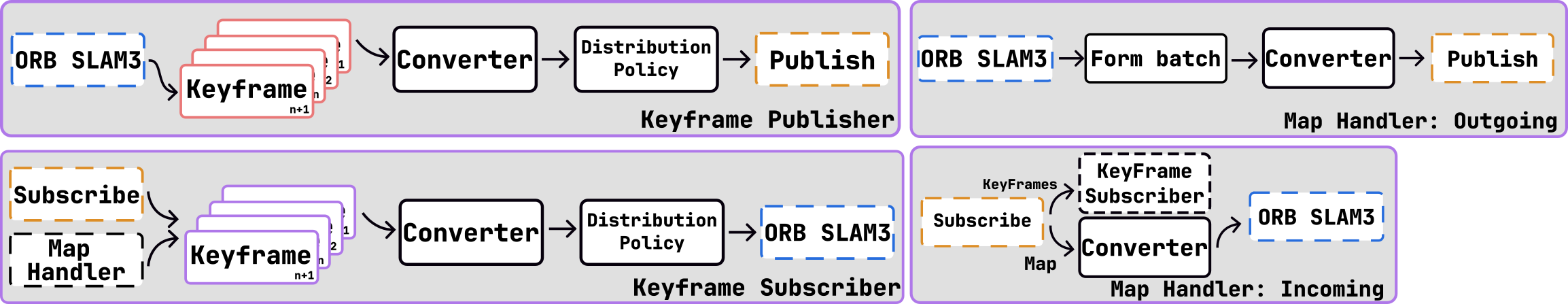}
  \caption{State managers for Keyframes and Maps. A dashed outline indicates that data is transmitted to another module/layer.}
  \label{fig:state_managers_all}
\end{figure*}







\subsection{Distribution layer} \label{subsect:dist_layer}

The distribution layer functions as an intermediary between the core and communication layers, with the responsibility of managing the overall state and updating the SLAM state accordingly. 
The overall state is stored in the observer module, which is managed by three state managers: KeyFrame publisher, Subscriber, and Map Handler. 
The system architecture and its internal modules can be seen in Figure~\ref{fig:sys_arch}, and all the state management pipelines can be seen in Figure~\ref{fig:state_managers_all}.

\textbf{The Observer} module is primarily responsible for storing the overall state and facilitating communication between sub-modules within the distribution layer and the SLAM system. 
It captures changes in the SLAM state when data propagates to the state managers, which, in turn, access the overall state in the observer when necessary.
Additionally, it can route KFs back to the SLAM system if no devices are discovered in the network.

\textbf{Keyframe Publisher and Subscriber} operate on separate threads to ensure real-time performance, as updating the nodes (and receiving updates) in the network with the latest data is critical.
Two types of keyframe interfaces are defined: new keyframes and keyframe updates. 
New keyframes provide the complete object definition, while keyframe updates include only the updated parts, such as pose, connections, and IDs of visible map points. 
Therefore, new keyframe messages are larger in size. 
Both types follow the same processing pipeline, from message receipt to integration into the SLAM system. 
New keyframes are published independently via their own topic for quick updates, whereas keyframe updates are bundled with map messages to avoid network congestion due to frequent updates.
Finally, keyframes can be assigned to three targets: no target (only inserted to the current map or update an existing keyframe), LM, or LC.

\textit{Incoming keyframes} are received by the communication layer and placed into a stack managed by the KeyFrame Subscriber, which processes them in a First-in-First-Out (FIFO) order. 
Keyframes are converted to ORB SLAM3 interface, including the associated map points. 
If a keyframe exists in the state, then it is updated with new parameters. 
Otherwise, the keyframe is added to the map.
Finally, the keyframe is forwarded to its target module following the distribution policy. 

\textit{Outgoing keyframes} follow a similar pipeline. They are placed into a stack and processed in a FIFO order. 
Keyframes are converted to ROS2 and distributed according to the distribution policy via the communication layer. 

\textbf{Map Handler} 
\label{subsect:map_handler}
operates on a single thread, handling incoming and outgoing maps. 
Two types of maps are defined: Local and global maps.
Both map types have similar publish/subscribe pipelines, with slightly different parameter selections.
The local map focuses on local bundle adjustments, while the global map encompasses broader updates, such as Map Merge (MM), LC, and GBA. 
When a global map update (GBA/LC/MM) is initiated, the other nodes are notified to halt the construction of new keyframes (TR) and the processing of the latest unoptimized keyframes (LM), following a strategy similar to that employed in ORB-SLAM3.
Execution is stopped to prevent data in two different coordinate frames since a global map update often tends to transform the complete trajectory. 
Execution remains stopped until the full global map is received on each node. 

\textit{Outgoing maps} are formed by keeping track of locally updated keyframes and map points via the Observer module. 
The most recent keyframe is retrieved from the stack with \(n\) covisible keyframes which are included in the outgoing map.
Maps are published in small batches to avoid network congestion and large message sizes.
The number of KFs increases in consecutive batches if no new map updates are performed in between.
Local map updates occur either after LBA or a set time interval \(t_{lmfreq}\) if unpublished updates exist. 
Global map updates occur after GBA/LC/MM has finished. 
Each batch is converted to ROS2 and distributed via the communication layer. 

\textit{Incoming maps} follow a similar pipeline as incoming keyframes.
Keyframes in the map are propagated to the KeyFrame subscriber, the map message is converted to ORB SLAM3, and an existing map is updated.

\begin{table*}
\smallskip
  \caption{Results of 1-node (1), and 3-node (3) configurations. All metrics are measured at TR node. Total bandwidth (BW) in Mbps and message frequencies (KF, Map) reported from network (NW) measurements. }
  \begin{subtable}[t]{\textwidth}
  \caption{EuRoC datasets}
  \resizebox{\textwidth}{!}{%
  \setlength{\tabcolsep}{0.55em} 
  {\renewcommand{\arraystretch}{1.01}
  \begin{tabular}{|c c|c c c c c | c c c | c c |}
    \hline
     & &  \multicolumn{10}{c|}{EuRoC} \\
     & &  MH\_01 & MH\_02 & MH\_03 & MH\_04\(^{g}\) & MH\_05\(^{g}\) & V1\_01\(^{g}\) & V1\_02\(^{g}\) & V1\_03\(^{g}\) & V2\_01\(^{g}\) & V2\_02\(^{g}\) \\
    \hline
    \multirow{3}{*}{\rotatebox[origin=c]{90}{ATE}} & ORB SLAM3 & 0.042 & 0.034 & 0.038 & 0.080 & 0.053 & 0.056 & 0.064 & 0.168 & 0.060 & 0.084\\
    & DSLAM (1) (ours) & \textbf{0.051} & \textbf{0.040} & \textbf{0.040} & 0.533 & \textbf{0.060} & 0.186 & \textbf{0.068} & 0.207 & \textbf{0.065} & \textbf{0.191} \\
    & DSLAM (3) (ours) & \textbf{0.042} & \textbf{0.036} & 0.176 & \textbf{0.067} & \textbf{0.073} & \textbf{0.053} & \textbf{0.084} & 0.208 & \textbf{0.061} & 0.153 \\
    \hline
    \multirow{3}{*}{\rotatebox[origin=c]{90}{Fails}} & ORB SLAM3  & 0.0 & 0.0 & 0.0 & 0.0 & 0.0 & 0.0 & 0 & 2.3 & 0.0 & 0.3 \\
    & DSLAM (1) (ours) & \textbf{0.0} & \textbf{0.0} & \textbf{0.0} & 0.0 & \textbf{0.0} & 0.33 & \textbf{0.0} & 0.67 & \textbf{1.0} & \textbf{0.67} \\
    & DSLAM (3) (ours) & \textbf{0.0} & \textbf{0.0} & 0.33 & \textbf{0.0} & \textbf{0.67} & \textbf{0.0} & \textbf{0.33} & 0.67 & \textbf{1.0} & 1.0 \\
    \hline
    \multirow{3}{*}{\rotatebox[origin=c]{90}{CPU}} & ORB SLAM3  & 50.06 & 50.59 & 50.07 & 50.62 & 50.30 & 51.32 & 51.87 & 51.59 & 52.39 & 51.59 \\
    & DSLAM (1) (ours) & \textbf{47.90} & \textbf{48.27} & \textbf{48.65} & 50.29 & \textbf{50.49} & 50.00 & \textbf{50.08} & 51.00 & \textbf{48.36} & \textbf{49.87} \\
    & DSLAM (3) (ours) & \textbf{46.23} & \textbf{47.64} & 47.14 & \textbf{47.85} & \textbf{49.10} & \textbf{48.17} & \textbf{49.79} & 48.63 & \textbf{48.86} & 50.36 \\
    \hline
    \multirow{2}{*}{\rotatebox[origin=c]{90}{NW}} & BW (TR/LM/LC) & 18/35/17 & 21/39/19 & 17/34/16 & 17/32/15 & 18/34/17 & 21/39/19 & 20/38/19 & 16/29/14 & 20/38/20 & 23/43/22 \\
    & Frequency (KF/Map) & 3.0/6.2 & 3.2/6.5 & 3.4/6.3 & 4.7/7.2 & 4.7/7.9 & 3.1/6.6 & 5.0/7.8 & 6.0/8.0 & 4.0/7.6 & 5.1/8.2 \\
    \hline
  \end{tabular}
  }
  }%
  \label{tab:euroc_results}
  \end{subtable}%

  \begin{subtable}[t]{\textwidth}

  \caption{TUM datasets and real-life experiments.}
  \resizebox{\textwidth}{!}{%
  \setlength{\tabcolsep}{0.4em} 
  {\renewcommand{\arraystretch}{1.01}
  \begin{tabular}{|c c|c c c c c c | c c c c c c c |}
    \hline
     & &  \multicolumn{6}{c|}{TUM} & \multicolumn{7}{c|}{Office} \\
     & &  room1 & room2 & room3\(^{g}\) & room4\(^{g}\) & room5 & room6  & 01\(^{g}\) & 02\(^{g}\) & 03\(^{g}\) & 04\(^{g}\) & 05\(^{g}\) & 06\(^{g}\) & 07\(^{g}\) \\
    \hline
    \multirow{3}{*}{\rotatebox[origin=c]{90}{ATE}} & ORB SLAM3 & 0.090 & 0.048 & 0.059 & 0.079 & 0.106 & 0.075 & 0.0 & 0.0 & 0.0 & 0.0 & 0.0 & 0.0 & 0.0 \\
    & DSLAM (1) & 0.239 & \textbf{0.046} & 0.083 & \textbf{0.081} & 0.211 & \textbf{0.075} & \textbf{0.038} & \textbf{0.099} & \textbf{0.199}* & \textbf{0.105} & \textbf{0.061}* & \textbf{0.102} & \textbf{0.282}* \\
    & DSLAM (3) & \textbf{0.079} & \textbf{0.053} & \textbf{0.073} & \textbf{0.073} & \textbf{0.083} & \textbf{0.075} & \textbf{0.073} & \textbf{0.124} & \textbf{0.147}* & \textbf{0.260} & \textbf{0.138} & \textbf{0.134} & \textbf{0.048} \\
    \hline
    \multirow{3}{*}{\rotatebox[origin=c]{90}{Fails}} & ORB SLAM3  & 0.0 & 0.0 & 0.0 & 0.0 & 0.0 & 0.0 & 0.0 & 2.0 & 0.0 & 0.0 & 1.0 & 2.0 & 0.0  \\
    & DSLAM (1) & 0.0 & \textbf{0.0} & 0.0 & \textbf{0.0} & 0.0 & \textbf{0.0} & \textbf{0.0} & \textbf{3.0} & \textbf{1.0}* & \textbf{0.0} & \textbf{3.0}* & \textbf{1.0} & \textbf{1.0}* \\
    & DSLAM (3) & \textbf{0.0} & \textbf{0.0} & \textbf{0.0} & \textbf{0.0} & \textbf{0.0} & \textbf{0.0} & \textbf{0.0} & \textbf{2.0} & \textbf{2.0}* & \textbf{0.0} & \textbf{1.0} & \textbf{2.0} & \textbf{0.0} \\
    \hline
    \multirow{3}{*}{\rotatebox[origin=c]{90}{CPU}} & ORB SLAM3 & 56.69 & 55.75 & 57.60 & 57.14 & 56.95 & 53.67 & 57.32 & 56.61 & 57.92 & 59.82 & 53.37 & 58.24 & 61.71 \\
    & DSLAM (1) & 54.55 & \textbf{50.32} & 53.91 & \textbf{54.85} & 54.04 & \textbf{52.63} & \textbf{53.59} & \textbf{51.98} & \textbf{52.46}* & \textbf{55.50} & \textbf{48.94}* & \textbf{51.21} & \textbf{54.56} \\
    & DSLAM (3) & \textbf{51.84} & \textbf{49.68} & \textbf{52.26} & \textbf{52.83} & \textbf{52.07} & \textbf{48.28} & \textbf{46.24} & \textbf{45.75} & \textbf{45.10}* & \textbf{45.72} & \textbf{45.92} & \textbf{44.90} & \textbf{44.91} \\
    \hline
    \multirow{2}{*}{\rotatebox[origin=c]{90}{NW}} & Bandwidth & 31/60/29 & 20/40/19 & 18/35/17 & 23/47/23 & 34/65/31 & 21/40/20 & 13/19/10 & 11/21/9 & 10/21/11 & 11/21/12 & 13/26/11 & 15/28/15 & 17/34/17 \\
    & Frequency & 3.2/6.7 & 2.0/4.6 & 3.1/6.8 & 3.5/6.6 & 3.6/7.2 & 2.1/4.9 & 3.7/7.2 & 3.0/5.6 & 3.6/6.7 & 3.3/6.7 & 3.2/6.1 & 3.1/6.3 & 3.7/7.1 \\

    \hline
  \end{tabular}
  }
  }%
  \label{tab:tum_results}%

  \end{subtable}
  \vspace{0.1mm} \\
  \footnotesize{$^1$ Experiments where Global map update (GBA/LC/MM) was performed at least once are denoted by \(^g\).}

  \footnotesize{$^2$ Experiments with one or more sequences having high ATE due to incorrect state management (max ATE \(\geq\) 1.0m) are not marked in bold.}
   
  \footnotesize{$^3$ Experiments which did not complete whole sequence (\(\leq\) 90\% succesful tracking) are denoted by *.}
  
\end{table*}

\subsection{Communication Layer}

The communication layer functions as a gateway between the distribution layer and the communication middleware. 
It propagates data from the network to the inner layers and vice versa. 
Minimal processing is performed in the communication layer to prevent congestion in the subscription threads. 
ROS2 was selected due to its popularity within the robotics research community.


\section{IMPLEMENTATION DETAILS} \label{experimental_details}

\subsection{Experiment setup} \label{subsec:setup}

The proposed framework is implemented in C++ with ROS2 Humble as the communication middleware, utilizing eProsima's Fast DDS. 
Monocular ORB SLAM3 \cite{orbslam3_campos_21} serves as the core SLAM system. 
The system is containerized using Docker, built for ARM64 and x86. 
ROS2 employs reliable QoS for all topics. 
Input images are read from rosbag and delivered via ROS2, except in real-life experiments where images are published from the camera via ROS2. 
The distribution layer is configured to send keyframe updates in batches; local maps start with three keyframes and increase to fifteen, with delays of 50 ms between updates. 
Global map batches consist of ten keyframes, sent every 100 ms. 
SLAM initialization follows the original ORB SLAM3 configuration \cite{orbslam3_campos_21}. 

The hardware setup is as follows: 
TR runs on NVIDIA Xavier (6-core ARM 64-bit CPU with 8 GB LPDDR4 memory), 
LM on Intel NUC 13 (Intel i5-13600K 14-core CPU and 64 GB DDR5 memory), and 
LC on NVIDIA Orin (8-core ARM 64-bit CPU and 16 GB LPDDR5 memory). 
NVIDIA Xavier and Orin are wirelessly connected to a router, whereas Intel NUC is using a wired connection. 
All computations are executed on the CPU, as ORB SLAM3 does not internally utilize the GPU.

Two types of experiments were conducted using benchmark datasets. 
In the first set of experiments, the proposed system is executed in one node configuration, reverting to standalone ORB SLAM3. 
The second set of experiments involved distributing all SLAM modules to three computing nodes.
Finally, real-world experiments were carried out using the latter setup, where an OAK-D Lite USB camera was used. 
A detailed description of the real-world experiments can be found in Section~\ref{subsect:rl_experiments}.

\subsection{Benchmark datasets: EuRoC and TUM}

The experiments were conducted using the EuRoC \cite{Burri_euroc_2016} and TUM \cite{klenk_tumvie_2021, schubert_vidataset_18} datasets. 
The EuRoC Machine Hall (MH) dataset simulates aerial robotics scenarios with a UAV operating in an industrial environment, while the Vicon Room 1 and 2 datasets represent augmented reality (AR) scenarios with more crowded spaces and rapid movements. 
All EuRoC datasets provide millimeter-accurate ground truth from a laser tracking system \cite{Burri_euroc_2016}. 
V2\_03 was excluded, as monocular ORB SLAM3 did not successfully complete all sequences, as reported in the original work \cite{orbslam3_campos_21}.
Rooms 1–6 from the TUM dataset \cite{klenk_tumvie_2021, schubert_vidataset_18}, featuring full motion capture trajectories as ground truth, were used.
All benchmark datasets were recorded in environments where direct communication with a router could be feasible.

\subsection{Real-life Experiments: Office} \label{subsect:rl_experiments}

Real-life experiments were conducted in an office environment to verify the correct distributed execution of SLAM. 
Experiments of varying difficulty were performed with varying camera movement speeds.
Challenging sequences aimed to induce TR failures and test the correct operation of MM. 
The confined space also facilitated multiple LCs and GBAs.

Rosbags were recorded from all experiments to evaluate performance under a single-node setup and the original ORB-SLAM3, which are unaffected by network conditions. 
During live experiments, distributed ORB-SLAM3 was executed with three nodes.
Seven experiments were conducted using the OAK-D Lite USB camera setup. The camera published RGB frames to a ROS2 topic at 20 fps with manual focus, exposure, and white balance settings. Frames were downscaled to \(640 \times 320\), and 1000 ORB features were extracted from each frame.

\subsection{Evaluation metrics}

We evaluated the system using the following metrics: 
\textbf{CPU Utilization} in percent; 
\textbf{RMS ATE} in meters for accuracy; 
\textbf{Failure rate}, times TR loses track;
\textbf{Network bandwidth} in Mbps; and 
\textbf{Frequency} of published KF and map messages.
The standard accuracy metric, RMS ATE \cite{slam_metrics_sturm_12}, is calculated using the evo Python package \cite{grupp_evo_2017} from the ground truth trajectory with scaling and alignment.
CPU utilization of the process is measured by reading the pseudo-filesystem /proc and through kernel calls. 
Network measurements are captured using tcpdump and analyzed with Wireshark, with network bandwidth representing the average bandwidth over the entire duration. 
We performed three experiments on each benchmark dataset and computed the average results. The same metrics were used for both real-life and benchmark experiments. 
However, in real-life experiments, the estimated trajectory from the original ORB-SLAM3 was utilized to assess the relative accuracy of the distributed system.

\section{RESULTS AND DISCUSSION}


\subsection{Benchmark datasets: EuRoC and TUM}

\subsubsection{3-node setup} 
\begin{figure}[tp]
  \centering
  \smallskip
    \includegraphics[width=\columnwidth]{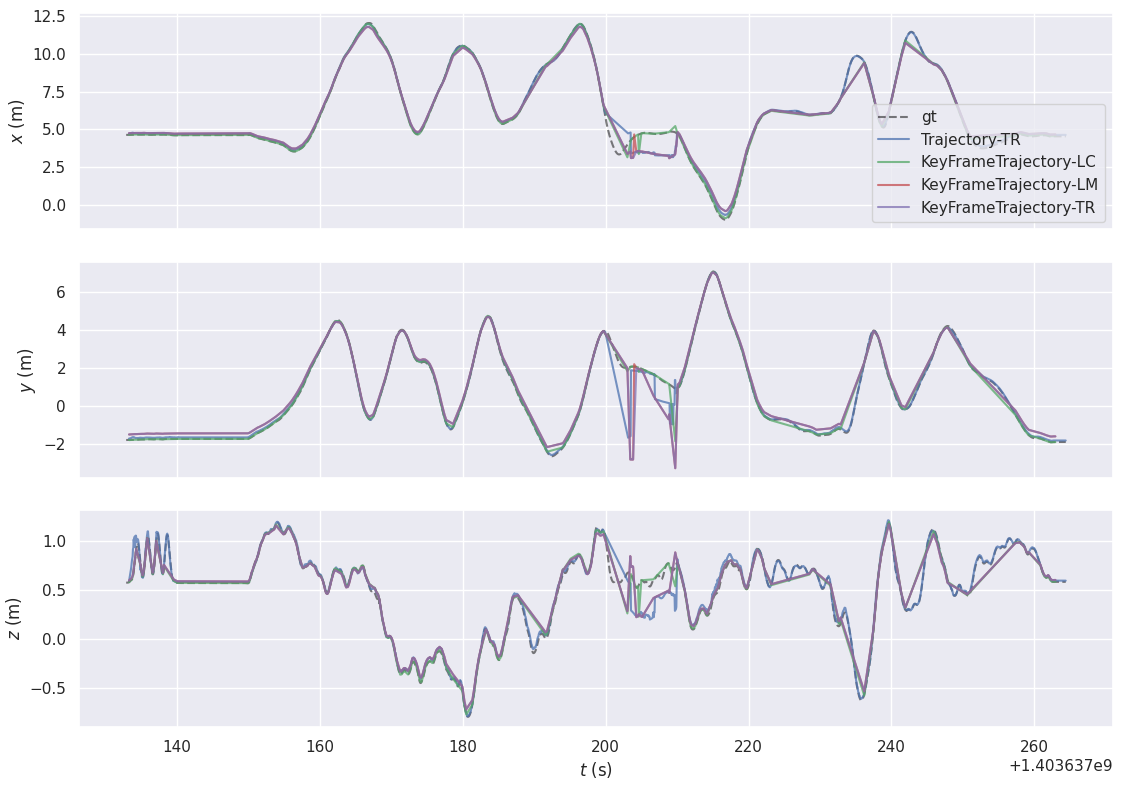}
  \caption{Artifacts can be seen between timesteps 200-220. KeyFrames and full trajectory plotted against ground truth. EuRoC, MH\_03 dataset.}
  \label{fig:artifacts}
\end{figure}

The fully distributed ORB SLAM3 achieves comparable performance to the original system across most of the experimented datasets, with an accuracy within \(\pm 0.00-0.02m\) when all sequences were executed without state management issues. 
The results are marked in bold in Tables~\ref{tab:euroc_results} and \ref{tab:tum_results}. 
Some experiments captured inconsistencies between nodes that introduced "artifacts" into the final trajectory estimation, as illustrated in Figure~\ref{fig:artifacts}. 
We could not identify a definitive reason for these artifacts, as we were unable to reproduce the issues in a deterministic manner, even on the same dataset. 
For instance, the RMSE ATE of all MH\_03 experiments were \((0.040, 0.818, 0.037)\), where the second experiment was the only one with poor accuracy. 
Similar issues were found in the EuRoC datasets MH\_03, V1\_03, and V2\_02 (not bold in Table~\ref{tab:euroc_results}), where one or two sequences exhibited higher ATE. 

Other metrics, such as the number of failures, also aligned with the original system, with minor increases in datasets such as EuRoC MH\_05 and V1\_02. 
However, multiple maps were merged together, yielding high accuracy in the resulting trajectory. 
The distribution of SLAM modules slightly decreases CPU utilization by \(1.2-6.1\%\), even with unoptimized processing and communication. 
We believe resource usage could be improved by optimizing the system, e.g. minimizing interface definitions, intelligent distribution policy, and optimizing interface conversions. 

Even in cases of failed state management, we consider that the distribution system can execute SLAM in a fully distributed setting with comparable performance to the original system since the failure cases are highly irregular. Possible reasons for the issues are discussed in Section~\ref{subsect:state_issues}.

\subsubsection{Network analysis} 
Bandwidth and message frequency measurements can be seen in the Tables~\ref{tab:euroc_results} and \ref{tab:tum_results}.
\(2.1-6.0\) KF and \(4.6-8.2\) map messages are published per second, depending on the dataset.
Total network usage is within \(9-31 Mbps\) on TR and LC nodes, and \(19-60 Mbps \) on LM node. LM has higher network usage because it updates local maps to both TR and LC nodes.
It can be seen that smaller environments (Room 1-6) require higher bandwidth than larger environments (MH\_01-05) due to high covisibility between keyframes, which increases the number of keyframes that LM optimizes in LBA. 
Network usage at LC varies depending on global map updates; however, they are still predominantly influenced by incoming transmissions from LM, similar to TR. 
The study is performed in a controlled local area network, not including variance of, e.g., the internet.

\subsubsection{1-node setup}
Results from the system reverting to the standalone ORB SLAM3 show similar outcomes. 
The RMSE ATE is within \(\pm 0.002-0.107 m\) compared to the ORB SLAM3, without the state management issues. 
Temporary failures in state management can be observed in datasets V1\_01, V1\_03, Room1, and 5. E.g, ATEs for V1\_01 are \((0.452, 0.054, 0.053)\).
Other metrics, i.e., number of failures and CPU usage, exhibit similar performance to the original system and the 3-node configuration.



\subsection{Real-life experiments: Office}

In real-life experiments, the 3-node distribution achieved ATE within \(0.048-0.260 m\) when scaled and aligned to the trajectory estimated by the ORB SLAM3. 
The experiments were designed to include multiple LC/MM/GBAs, which is reflected in the increased number of failures compared to the benchmark datasets, as shown in Table~\ref{tab:tum_results}. 
In most cases, the system successfully merged multiple maps (office02, 06); however, in office03, the system was able to merge only 2 out of 3 maps. 
Additionally, multiple LC/GBAs were performed successfully in office03 and office04. 
Similar results can be observed in the 1-node setup (office03, 05), where multiple map merges were completed successfully. 
Furthermore, similar to the 3-node setup, the 1-node setup achieved a comparable ATE of  \((0.038-0.282 m)\).
CPU and network metrics show similar performance to the benchmark datasets.

\subsection{Discussion: Issues with state management}
\label{subsect:state_issues}
We identified two common themes surrounding the issues related to state management: map initialization and global map updates. 
Modifications to the initialization (Section~\ref{subsect:core}) can sometimes allow tracking to continue, even in cases of poor initial KF pose estimation (by TR or LM).
Initialization was more likely the cause of the issues with datasets Room1 and Room5 in 1-node configuration. 
On the other hand, artifacts were sometimes present when a global map update was performed. 
Identifying a definitive cause is complicated since the issues are not present in the majority of the experiments where a global map update was performed.
Our theory is that signal to pause other nodes when the global map update has started is delayed or lost, which can produce unaligned KFs compared to the optimized global map.
Investigation of the internal (map initialization and GBA) and external conditions (network conditions) to the state management is left as future work.
\section{CONCLUSION}

In this article, we have proposed a novel self-organizing distribution framework for VSLAM that can allocate SLAM modules across a network of devices.
We have implemented and demonstrated the distribution framework for monocular ORB SLAM3 and empirically evaluated its performance against the original system using benchmark datasets and real-life experiments. 
The results demonstrate that the novel distribution framework is capable of fully distributing all SLAM modules across a network of heterogeneous devices and self-orchestrating collaborative execution with comparable accuracy and resource utilization to the original system. 
Additionally, we have shown that the system can degrade into standalone ORB SLAM3 in single-node configuration, e.g., when connectivity to other nodes is lost. 
Finally, we have identified several current limitations and proposed new future research directions.


\bibliographystyle{IEEEtran.bst} 
\bibliography{references.bib}

\begin{thebibliography}{10}
\providecommand{\url}[1]{#1}
\csname url@rmstyle\endcsname
\providecommand{\newblock}{\relax}
\providecommand{\bibinfo}[2]{#2}
\providecommand\BIBentrySTDinterwordspacing{\spaceskip=0pt\relax}
\providecommand\BIBentryALTinterwordstretchfactor{4}
\providecommand\BIBentryALTinterwordspacing{\spaceskip=\fontdimen2\font plus
\BIBentryALTinterwordstretchfactor\fontdimen3\font minus \fontdimen4\font\relax}
\providecommand\BIBforeignlanguage[2]{{%
\expandafter\ifx\csname l@#1\endcsname\relax
\typeout{** WARNING: IEEEtran.bst: No hyphenation pattern has been}%
\typeout{** loaded for the language `#1'. Using the pattern for}%
\typeout{** the default language instead.}%
\else
\language=\csname l@#1\endcsname
\fi
#2}}

\bibitem{covins_schmuck_21}
P.~Schmuck, T.~Ziegler, M.~Karrer, J.~Perraudin, and M.~Chli, ``{COVINS:} visual-inertial {SLAM} for centralized collaboration,'' \emph{CoRR}, vol. abs/2108.05756, 2021.

\bibitem{ccslam_schmuck_19}
P.~Schmuck and M.~Chli, ``Ccm-slam: Robust and efficient centralized collaborative monocular simultaneous localization and mapping for robotic teams,'' \emph{Journal of Field Robotics}, vol.~36, no.~4, pp. 763--781, 2019.

\bibitem{dcslam_zhang_18}
H.~Zhang, X.~Chen, H.~Lu, and J.~Xiao, ``Distributed and collaborative monocular simultaneous localization and mapping for multi-robot systems in large-scale environments,'' \emph{International Journal of Advanced Robotic Systems}, vol.~15, no.~3, p. 1729881418780178, 2018.

\bibitem{slam_midterm_burgard_08}
W.~Burgard, O.~Brock, and C.~Stachniss, \emph{Mapping Large Loops with a Single Hand-Held Camera}, 2008, pp. 297--304.

\bibitem{orbslam3_campos_21}
C.~Campos, R.~Elvira, J.~J.~G. Rodríguez, J.~M. M.~Montiel, and J.~D.~Tardós, ``Orb-slam3: An accurate open-source library for visual, visual–inertial, and multimap slam,'' \emph{IEEE Transactions on Robotics}, vol.~37, no.~6, pp. 1874--1890, 2021.

\bibitem{vinsmono_li_18}
T.~Qin, P.~Li, and S.~Shen, ``Vins-mono: A robust and versatile monocular visual-inertial state estimator,'' \emph{IEEE Transactions on Robotics}, vol.~34, no.~4, pp. 1004--1020, 2018.

\bibitem{lif_slam_dl_bruno_21}
H.~M.~S. Bruno and E.~L. Colombini, ``Lift-slam: A deep-learning feature-based monocular visual slam method,'' \emph{Neurocomputing}, vol. 455, pp. 97--110, 2021.

\bibitem{ali_edgeslam_22}
A.~J. Ben~Ali, M.~Kouroshli, S.~Semenova, Z.~S. Hashemifar, S.~Y. Ko, and K.~Dantu, ``Edge-slam: Edge-assisted visual simultaneous localization and mapping,'' \emph{ACM Trans. Embed. Comput. Syst.}, vol.~22, no.~1, oct 2022.

\bibitem{swarmmap_xu_22}
J.~Xu, H.~Cao, Z.~Yang, L.~Shangguan, J.~Zhang, X.~He, and Y.~Liu, ``$\{$SwarmMap$\}$: Scaling up real-time collaborative visual $\{$SLAM$\}$ at the edge,'' in \emph{19th USENIX Symposium on Networked Systems Design and Implementation (NSDI 22)}, 2022, pp. 977--993.

\bibitem{edgeslam2_li_24}
D.~Li, Y.~Zhao, J.~Xu, S.~Zhang, L.~Shangguan, and Z.~Yang, ``edgeslam2: Rethinking edge-assisted visual slam with on-chip intelligence,'' in \emph{IEEE INFOCOM 2024 - IEEE Conference on Computer Communications}, 2024, pp. 1481--1490.

\bibitem{edgeslam_semantic_cao_23}
H.~Cao, J.~Xu, D.~Li, L.~Shangguan, Y.~Liu, and Z.~Yang, ``Edge assisted mobile semantic visual slam,'' \emph{IEEE Transactions on Mobile Computing}, vol.~22, no.~12, pp. 6985--6999, 2023.

\bibitem{dynnetslam_sossala_22}
P.~Sossalla, J.~Hofer, J.~Rischke, C.~Vielhaus, G.~T. Nguyen, M.~Reisslein, and F.~H.~P. Fitzek, ``Dynnetslam: Dynamic visual slam network offloading,'' \emph{IEEE Access}, vol.~10, pp. 116\,014--116\,030, 2022.

\bibitem{cloudslam_wright_20}
K.-L. Wright, A.~Sivakumar, P.~Steenkiste, B.~Yu, and F.~Bai, ``Cloudslam: Edge offloading of stateful vehicular applications,'' in \emph{2020 IEEE/ACM Symposium on Edge Computing (SEC)}, 2020, pp. 139--151.

\bibitem{edge_slam_energy_sossala_23}
P.~Sossalla, J.~Hofer, C.~Vielhaus, J.~Rischke, and F.~H.~P. Fitzek, ``Offloading visual slam processing to the edge: An energy perspective,'' in \emph{2023 International Conference on Information Networking (ICOIN)}, 2023, pp. 39--44.

\bibitem{seminal_slam_lemmer_88}
R.~Smith, M.~Self, and P.~Cheeseman, ``Estimating uncertain spatial relationships in robotics,'' in \emph{Uncertainty in Artificial Intelligence}, ser. Machine Intelligence and Pattern Recognition, J.~F. LEMMER and L.~N. KANAL, Eds.\hskip 1em plus 0.5em minus 0.4em\relax North-Holland, 1988, vol.~5, pp. 435--461.

\bibitem{seminal_slam_smith_86}
R.~C. Smith and P.~Cheeseman, ``On the representation and estimation of spatial uncertainty,'' \emph{The International Journal of Robotics Research}, vol.~5, no.~4, pp. 56--68, 1986.

\bibitem{covinsg_patel_23}
M.~Patel, M.~Karrer, P.~Bänninger, and M.~Chli, ``Covins-g: A generic back-end for collaborative visual-inertial slam,'' 2023.

\bibitem{monoslam_davison_03_1}
Davison, ``Real-time simultaneous localisation and mapping with a single camera,'' in \emph{Proceedings Ninth IEEE International Conference on Computer Vision}, 2003, pp. 1403--1410 vol.2.

\bibitem{monoslam_davison_07_2}
A.~J. Davison, I.~D. Reid, N.~D. Molton, and O.~Stasse, ``Monoslam: Real-time single camera slam,'' \emph{IEEE Transactions on Pattern Analysis and Machine Intelligence}, vol.~29, no.~6, pp. 1052--1067, 2007.

\bibitem{monoslam_civera_08_3}
J.~Civera, A.~J. Davison, and J.~M.~M. Montiel, ``Inverse depth parametrization for monocular slam,'' \emph{IEEE Transactions on Robotics}, vol.~24, no.~5, pp. 932--945, 2008.

\bibitem{orbslam1_murartal_15}
R.~Mur-Artal, J.~M.~M. Montiel, and J.~D. Tardós, ``Orb-slam: A versatile and accurate monocular slam system,'' \emph{IEEE Transactions on Robotics}, vol.~31, no.~5, pp. 1147--1163, 2015.

\bibitem{orbslam2_murartal_16}
R.~Mur-Artal and J.~Tardos, ``Orb-slam2: an open-source slam system for monocular, stereo and rgb-d cameras,'' \emph{IEEE Transactions on Robotics}, vol.~PP, 10 2016.

\bibitem{cudasift_slam_tardos_24}
R.~Elvira, J.~Tardós, and J.~Montiel, ``Cudasift-slam: multiple-map visual slam for full procedure mapping in real human endoscopy,'' 05 2024.

\bibitem{ransac_civera_10}
J.~Civera, O.~Grasa, A.~Davison, and J.~Montiel, ``1-point ransac for extended kalman filtering: Application to real-time structure from motion and visual odometry,'' \emph{J. Field Robotics}, vol.~27, pp. 609--631, 09 2010.

\bibitem{droid_teed_21}
Z.~Teed and J.~Deng, ``Droid-slam: Deep visual slam for monocular, stereo, and rgb-d cameras,'' \emph{Advances in neural information processing systems}, vol.~34, pp. 16\,558--16\,569, 2021.

\bibitem{dbow_galvez_12}
D.~Galvez-López and J.~D. Tardos, ``Bags of binary words for fast place recognition in image sequences,'' \emph{IEEE Transactions on Robotics}, vol.~28, no.~5, pp. 1188--1197, 2012.

\bibitem{ptam_klein_07}
G.~Klein and D.~Murray, ``Parallel tracking and mapping for small ar workspaces,'' in \emph{2007 6th IEEE and ACM International Symposium on Mixed and Augmented Reality}, 2007, pp. 225--234.

\bibitem{dgs_choudhary_17}
S.~Choudhary, L.~Carlone, C.~Nieto, J.~Rogers, H.~I. Christensen, and F.~Dellaert, ``Distributed mapping with privacy and communication constraints: Lightweight algorithms and object-based models,'' \emph{The International Journal of Robotics Research}, vol.~36, no.~12, pp. 1286--1311, 2017.

\bibitem{adapt_slam_chen_23}
Y.~Chen, H.~Inaltekin, and M.~Gorlatova, ``{AdaptSLAM}: Edge-assisted adaptive slam with resource constraints via uncertainty minimization,'' in \emph{Proc. IEEE INFOCOM}, 2023.

\bibitem{distributed_steen_2023}
M.~v. Steen and A.~S. Tanenbaum, \emph{\BIBforeignlanguage{en}{Distributed {Systems}}}, fourth edition, version 4.01 (january 2023)~ed.\hskip 1em plus 0.5em minus 0.4em\relax Erscheinungsort nicht ermittelbar: Maarten van Steen, 2023.

\bibitem{Burri_euroc_2016}
M.~Burri, J.~Nikolic, P.~Gohl, T.~Schneider, J.~Rehder, S.~Omari, M.~W. Achtelik, and R.~Siegwart, ``The euroc micro aerial vehicle datasets,'' \emph{The International Journal of Robotics Research}, 2016.

\bibitem{klenk_tumvie_2021}
S.~Klenk, J.~Chui, N.~Demmel, and D.~Cremers, ``Tum-vie: The tum stereo visual-inertial event dataset,'' in \emph{International Conference on Intelligent Robots and Systems (IROS)}, 2021.

\bibitem{schubert_vidataset_18}
D.~Schubert, T.~Goll, N.~Demmel, V.~Usenko, J.~Stueckler, and D.~Cremers, ``The tum vi benchmark for evaluating visual-inertial odometry,'' in \emph{International Conference on Intelligent Robots and Systems (IROS)}, October 2018.

\bibitem{slam_metrics_sturm_12}
J.~Sturm, N.~Engelhard, F.~Endres, W.~Burgard, and D.~Cremers, ``A benchmark for the evaluation of rgb-d slam systems,'' in \emph{2012 IEEE/RSJ International Conference on Intelligent Robots and Systems}, 2012, pp. 573--580.

\bibitem{grupp_evo_2017}
M.~Grupp, ``evo: Python package for the evaluation of odometry and slam.'' \url{https://github.com/MichaelGrupp/evo}, 2017.

\end{thebibliography}

\addtolength{\textheight}{-12cm}   





\end{document}